\documentclass[10pt,twocolumn,letterpaper]{article}

\usepackage{wacv}
\usepackage{times}
\usepackage{epsfig}
\usepackage{graphicx}
\usepackage{amsmath}
\usepackage{amssymb}
\usepackage{multirow}
\usepackage{color,soul}
\usepackage{algorithm}
\usepackage{algpseudocode}
\usepackage{enumerate}
\usepackage{makecell}
\usepackage{pifont}
\newcommand{\cmark}{\ding{51}}%
\newcommand{\xmark}{\ding{55}}%

\def\wacvPaperID{1031} 

\ifwacvfinal
\def\assignedStartPage{1} 
\fi

\ifwacvfinal
\usepackage[breaklinks=true,bookmarks=false]{hyperref}
\else
\usepackage[pagebackref=true,breaklinks=true,colorlinks,bookmarks=false]{hyperref}
\fi

\ifwacvfinal
\else
\fi

\begin{document}

\title{Joint Learning Architecture for Multiple Object Tracking and Trajectory Forecasting}

\author{Oluwafunmilola Kesa, Olly Styles, Victor Sanchez\\
University of Warwick\\
{\tt\small funmi.kesa | o.c.styles | v.f.sanchez-silva}
}

\maketitle

\begin{abstract}
This paper introduces a joint learning architecture (JLA) for multiple object tracking (MOT) and trajectory forecasting in which the goal is to predict objects' current and future trajectories simultaneously. Motion prediction is widely used in several state of the art MOT methods to refine predictions in the form of bounding boxes. 
Typically, a Kalman Filter provides short-term estimations to help trackers correctly predict objects' locations in the current frame. However, the Kalman Filter-based approaches cannot predict non-linear trajectories. We propose to jointly train a tracking and trajectory forecasting model and use the predicted trajectory forecasts for short-term motion estimates in lieu of linear motion prediction methods such as the Kalman Filter. We evaluate our JLA on the MOTChallenge benchmark. Evaluations result show that JLA performs better for short term motion prediction and reduces ID switches by $33\%$, $31\%$, and $47\%$ in the MOT16, MOT17, and MOT20 datasets, respectively, in comparison to FairMOT.

\end{abstract}


\section{Introduction}
Multiple object tracking (MOT) and trajectory forecasting are two different tasks in computer vision.
MOT aims to track the location of multiple objects in a video \cite{Pi2020JointlyInformation, Zhang2020FairMOT:Tracking, Wang2020TowardsTracking, Zhang2020}. It is a critical component for several applications, including autonomous driving \cite{desire}, video surveillance \cite{Lin2021OnTracking} and smart elderly care \cite{4809540}. Meanwhile, trajectory forecasting aims to predict future short- and long-term locations of objects in a video using the objects' past location information \cite{trajectory-forecasting-survey}.
Researchers have studied tracking and trajectory forecasting independently, and learning-based models exist for these two tasks.  However, the two tasks share similar properties as both require the objects of interest to be detected and re-identified across frames.

A recent formulation of trajectory forecasting called Multiple Object Forecasting (MOF) extends the traditional MOT to predict objects' future coordinates and scale in terms of their bounding boxes \cite{Styles2020MultipleEnvironments}. While MOF exploits the advantages of an object-based architecture, MOF requires pre-computed trajectories from an object tracker to estimate future locations. Separating tracking and trajectory forecasting poses some challenges in MOF, including high computational cost, that can limit its real-time application.

In MOT methods, issues such as identity (ID) switches and incorrect predictions are still dominant.  To this end, motion prediction is used to rectify wrong estimations 
caused by similar appearance embeddings or occlusion \cite{Wang2020TowardsTracking, Zhang2020FairMOT:Tracking}. More concretely, a Kalman Filter is typically used to provide short-term motion estimations, which are used to refine the bounding box estimations \cite{Wojke2018SimpleMetric, Zhang2020FairMOT:Tracking}. However, such a motion prediction method cannot predict non-linear trajectories. In contrast, trajectory forecasting can provide non-linear predictions that can benefit MOT methods.

Motivated by the above observations, we introduce the multiple object tracking and forecasting (MOTF) task,  whose aim is to simultaneously detect, track, and predict objects' current and future trajectories. We implement the MOTF task through a joint learning architecture (JLA) that models 
non-linear trajectories by using trajectory forecasts generated by an embedded forecasting network. By using these trajectory forecasts to refine bounding box estimations, our JLA can predict objects' locations during occlusion, an important advantage over previous works that rely on linear motion predictions. As a result, JLA can reduce ID switches substantially. 
Our contributions can be summarized as follows:
\begin{itemize}
    \item We introduce MOTF, a new task for multiple object tracking and trajectory forecasting (Section \ref{sec:motf}).
    \item We propose a novel architecture that jointly performs multiple object tracking and trajectory forecasting (Section \ref{sec:jla}).
    \item We employ the trajectory forecasts to refine objects' locations in lieu of the Kalman Filter.
    The trajectory forecasting model reduces the search space for an object’s location, thus, reducing the ID switches caused by similar appearance embeddings (Section \ref{sec:alg1}). 
    \item We introduce a new algorithm for data association of objects during occlusion by using the previous trajectory forecasts to estimate the location of occluded objects in subsequent frames (Section \ref{sec:alg2}). 
    \item We evaluate our model on the MOTChallange benchmarks. Our proposed method reduces ID switches 
    on the MOT20 benchmark by 47\% compared to FairMOT \cite{Zhang2020FairMOT:Tracking} (Section \ref{sec:result}).
\end{itemize}

\section{Related Work}

In this section, we review existing works on MOT and trajectory forecasting.

\subsection{Multiple Object Tracking}
Recent MOT methods leverage deep neural networks' representational power to learn the identity, appearance and pose of several objects to associate targets across several frames. Several of these methods follow the tracking-by-detection paradigm, in which objects are first detected as targets and then associated with subsequent detections, in the form  of the bounding boxes, 
to form trajectories \cite{Wojke2018SimpleMetric, Bewley2016SimpleTracking, Xiao2017JointSearch, Zhang2020FairMOT:Tracking, Pi2020JointlyInformation}. Specifically, these methods use bounding box estimations from an external detector and focus on improving the association of these estimations to form trajectories. Clearly, this approach can benefit from a strong object detector and a re-identification (re-ID) method. However, the high computational cost of  training an object detector and a re-ID model separately, and their slow inference time, limit the real-time application of such an approach. To address this issue, one approach is to train the object detector and  re-ID model simultaneously in an end-to-end manner. The works in \cite{Voigtlaender2019Mots:Segmentation, Wang2020TowardsTracking, Zhang2020FairMOT:Tracking} show that it is possible to design models that can simultaneously detect and predict identity embeddings by adding a re-ID head to existing object detectors. Our work takes this approach a step further to simultaneously detect, track, and forecast objects' locations by adding a trajectory forecast head to a tracking method. 

An important part of an MOT method is the task of association, which can be performed in an online or offline manner. Online methods \cite{Wojke2018SimpleMetric, Sun2021DeepTracking, Zhang2020FairMOT:Tracking, Bergmann2019TrackingWhistles, Pang2020Tubetk:Model, Wang2020TowardsTracking, Peng2020Chained-Tracker:Tracking} associate bounding box estimations sequentially up to the current frame. In offline methods \cite{choi2010multiple, 7410890, qin2012improving, yang2011learning, 8015000, henriques2011globally}, the order of association does not apply and future frame estimations can be used in data association. Offline methods can interpolate missing objects' locations by using the past, current and future estimations to generate better trajectory predictions than online methods. However, offline methods cannot be used in real-time applications \cite{LUO2021103448}. Our method uses trajectory forecasts to estimate occluded objects' locations in an online fashion.

\subsection{Trajectory Forecasting}
Trajectory forecasting methods predict the future location of an object identified in a video. Datasets for trajectory forecasting typically consist of footage of pedestrians or vehicles captured from a birds-eye perspective \cite{trajnet}. Methods use past location information in addition to various categories of features such as the interactions between objects \cite{Alahi2016SocialSpaces,learning-etiquette} and the estimated final destination \cite{kartik-endpoints,grid-based-goals}. A smaller number of works have also considered trajectory forecasting from an egocentric viewpoint \cite{fpl,brian-egocentric,Styles2020MultipleEnvironments}, where visual information such as human pose estimations or optical flow can be incorporated more easily. All the previous works mentioned use either ground truth or pre-computed object tracking results prior to forecasting.


\begin{figure*}[t]
    \centering
    \includegraphics[width=0.95\linewidth]{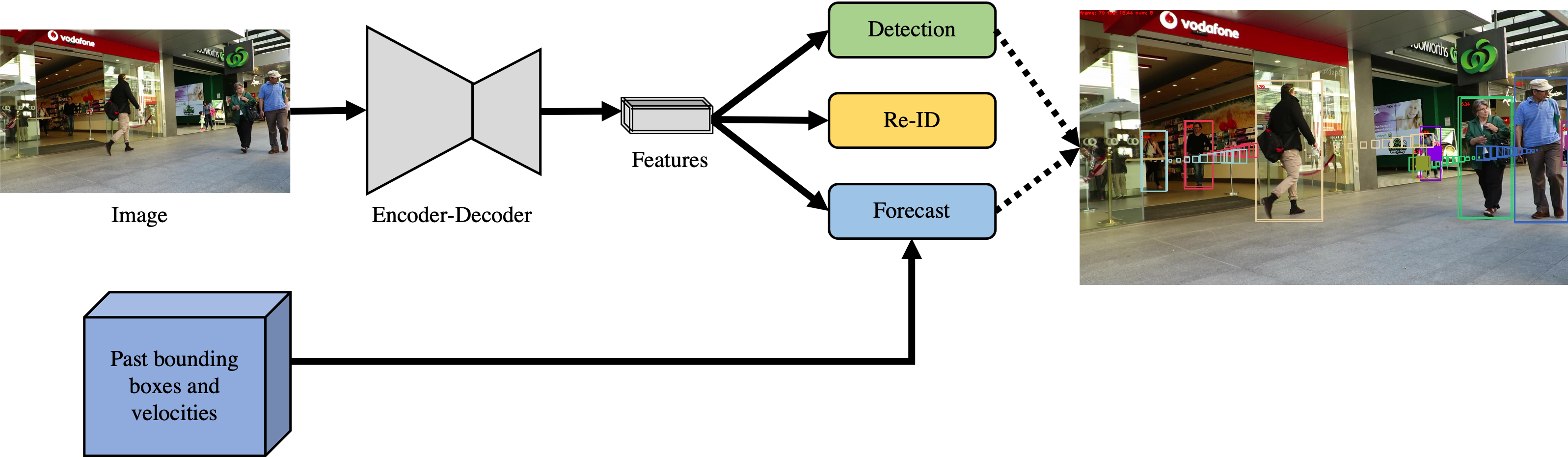}
    \caption{\textbf{JLA:} We propose a joint learning architecture for tracking  multiple objects and forecasting their trajectories. The trajectory forecasting branch takes the past bounding boxes and velocities, and an embedding as input to predict the  objects' locations into the future. JLA performs three tasks: detection, re-ID, and trajectory forecasting. The network is trained end-to-end.
    }
    \label{fig:jlaoverview}
\end{figure*}
\section{Multiple Object Tracking and Forecasting} \label{sec:motf}
MOTF draws from MOT and MOF and follows a similar formulation to these two tasks. In this section, we formalize the problem and explain the evaluation metrics.

\subsection{Problem Formulation} \label{sec:prob}
Consider a video with ${n}$ frames ${f_0, f_1, \ldots, f_{n-1}}$. Given frame $f_s$ at timestep $s$,
\begin{itemize}
    \item let $I$ be a set of identifiable objects in the frame such that $i \in I$, and
    \item let $\boldsymbol{b}_s = \{b^i_s\}_{i=1}^I$ be the set of bounding boxes for each identifiable object $i$ in frame $f_s$. Each bounding box $b^i_s=(x,y,w,h)$ is represented by the location of its centroid, $(x, y)$, and its width and height, $(w, h)$.
\end{itemize}
The aim of MOT is to associate all the frame-wise bounding boxes, $\{b^i_0\}, \{b^i_1\}, \ldots, \{b^i_{n-1}\}$ for all $i \in I$, to a unique identifier, $k \in 1,2 \dots K$, where $K$ is the total number of unique objects across all the frames, such that a set of tracks, $\mathcal{T} = \{t^k\}_{k=1}^K$, is computed for the entire video sequence, where $t^k$ represents the $k^{th}$ unique track. 
This association task can be formulated as a bipartite or linear assignment problem where only one bounding box is linked to another bounding box in a subsequent frame. Therefore, an object in frame $f_s$ is not associated with any other  object in the same frame.

Similarly, given a sequence of frames $f_{s-p}, f_{s-p+1}, \ldots, f_s$, with their respective set of tracks $\{t^k_{s-p}\},  \{t^k_{s-p+1}\}, \ldots, \{t^k_{s}\}$ for all $k \in K$, the task of MOF is to predict the future set of bounding boxes $\{b^k_{s+1}\}, \{b^k_{s+2}\}, \ldots, \{b^k_{s+q}\}$ for all $k \in K$, for future frames $f_{s+1}, f_{s+2}, \ldots, f_{s+q}$, where $p$ is the number of past frames used as input and $q$ is the length of predictions into the future \cite{Styles2020MultipleEnvironments}. In MOF, the tracks are pre-determined before forecasting. 

Our goal is to create a JLA that jointly tracks and forecasts objects' locations. Thus, we formulate the MOTF task as a joint problem of MOT and MOF. Given frame $f_s$ and a sequence of $p$ past bounding boxes $\{b^k_{s-p}\},  \{b^k_{s-p+1}\}, \ldots, \{b^k_{s-1}\}$ for all $k \in K$, MOTF aims to compute tracks $\{t^k_{s}\}$ for all $k \in K$ at frame $f_s$  and forecast each track's current and future bounding boxes, \ie, $\{b^k_{s}\}, \{b^k_{s+1}\}, \{b^k_{s+2}\}, \ldots, \{b^k_{s+q}\}$ for all $k \in  K$. Unlike MOF, the current frame bounding boxes, \ie, $\{b^k_{s}\}$ for all $k \in  K$, are also predicted to evaluate the accuracy of the jointly trained forecasting model in predicting the location of the detected objects. In this work, we set $p=10$ and $q=60$, which correspond to less than 1 second in the past and predicting 2 seconds into the future, respectively, at 30Hz.

\subsection{Evaluation Metrics} \label{sec: metrics}
MOTF is a joint task of tracking and forecasting. We employ the evaluation metrics of both tracking and forecasting. We use the \emph{CLEAR} metrics \cite{Bernardin2008EvaluatingMetrics}, and \emph{IDF1} values \cite{Ristani2016PerformanceTracking} to evaluate the trajectory tracking performance. We use \emph{ADE/FDE} \cite{Alahi2016SocialSpaces}, and \emph{AIOU/FIOU} \cite{Styles2020MultipleEnvironments} metrics to evaluate the trajectory forecasting performance. 


\begin{figure*}[t]
    \centering
    \includegraphics[width=0.95\linewidth]{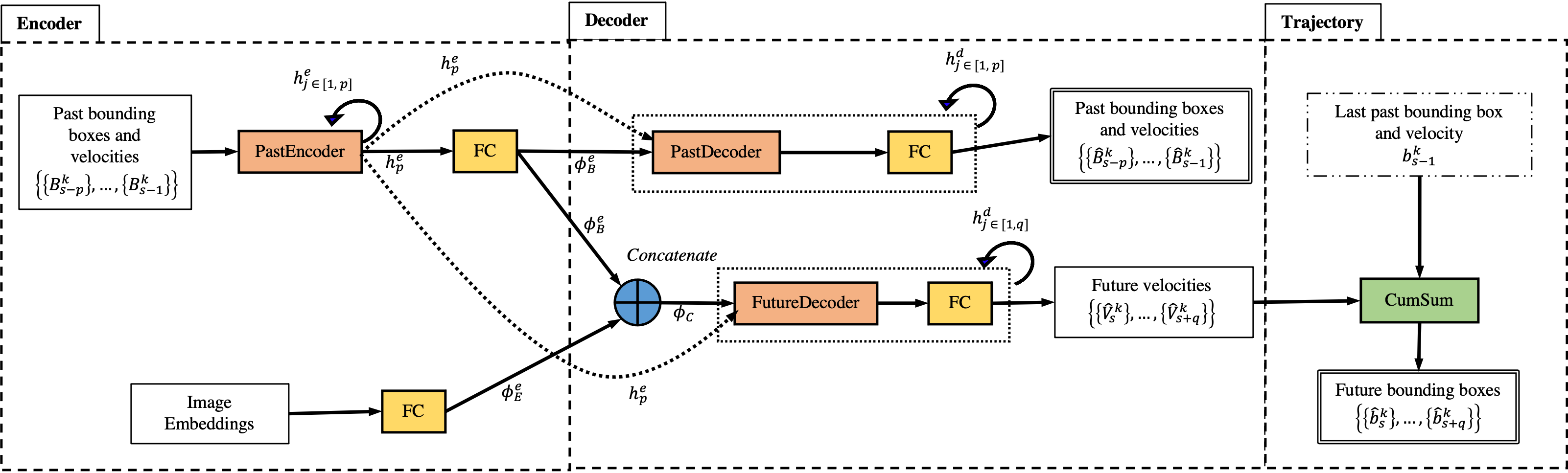}
    \caption{\textbf{Trajectory Forecast Architecture:} The trajectory  forecasting branch of JLA takes past bounding boxes, velocities, and DLA-34 feature embeddings (image embeddings) as input. It uses a Recurrent Neural Network (RNN)
    and a fully connected (FC) ReLU layer to encode the past bounding boxes, and a FC ReLU layer to encode the image embeddings. Two different decoders are used to decode past bounding boxes and velocities, and predict future velocities. Finally, a trajectory concatenation layer transforms the predicted future velocities into future bounding boxes.
    }
    \label{fig:jlaforecast}
\end{figure*}
\section{Proposed JLA} \label{sec:jla}
In this section, we present JLA, a joint learning architecture  
for MOTF. JLA draws from existing architectures for tracking and forecasting \cite{Zhang2020FairMOT:Tracking, Styles2020MultipleEnvironments, Ansari2020SimpleCpu}. We use the FairMOT model \cite{Zhang2020FairMOT:Tracking} as our base model because this architecture already performs detection and tracking. We add a forecasting branch to the network shown in Figure \ref{fig:jlaoverview}, and train the architecture end-to-end. FairMOT consist of a backbone network called DLA-34, an object detection head, and a re-ID head. More details about the FairMOT architecture can be found in  \cite{Zhang2020FairMOT:Tracking}.

The design of the trajectory forecasting branch is shown in Figure~\ref{fig:jlaforecast}. The goal of the this branch is to predict future bounding boxes of objects using the past bounding box information. The trajectory forecasting network consists of recurrent neural networks (RNN) used to encode and decode the past bounding boxes and predict future bounding boxes. The components of the network are listed below:
\begin{enumerate}[(i)]
\item An RNN to encode past bounding boxes and velocities $\{B^k_{s-p}\}, \ldots, \{B^k_{s-1}\}$, for all $k \in K$.
\item A fully-connected layer to encode DLA-34 feature embeddings retrieved from the tracking network.
\item An RNN to decode past bounding boxes and velocities.
\item An RNN to decode future velocities $\{\hat{V}^k_{s}\}, \ldots, \{\hat{V}^k_{s+q}\}$, for all $k \in K$.
\item A trajectory concatenation layer to convert future velocities to bounding boxes.
\end{enumerate}

\subsection{Past Bounding Box and Velocity Encoder}
An RNN (PastEncoder in Figure \ref{fig:jlaforecast}) is used to extract features from past bounding boxes. This encoder captures the velocity of each object by iterating over historical information. 

Given frame $f_s$  and past bounding boxes $\{b^k_{s-p}\},  \{b^k_{s-p+1}\}, \ldots, \{b^k_{s-1}\}$ for all $k \in K$, 
we construct a sequence of $p$ sets of 8-dimensional vectors $\boldsymbol{B} \in \mathbb{R}^{p \times 8}$. This sequence of 8-dimensional vectors can be written as $\boldsymbol{B} \equiv \{\{B^k_j\}^K_{k=1}\}^{s-1}_{j=s-p} \equiv \{\{B^k_{s-p}\}, \{B^k_{s-p+1}\}, \ldots, \{B^k_{s-1}\}\}$, where for each object $k$ in frame $f_j$, $B^k_j=(x^k_j, y^k_j, w^k_j, h^k_j, \Delta{x^k_j}, \Delta{y^k_j}, \Delta{w^k_j}, \Delta{h^k_j})$ and $(x^k_j, y^k_j)$ represents the location of the centroid of the corresponding bounding box, $(w^k_j, h^k_j)$ represents the width and height of the bounding box, $V^k_j = (\Delta{x^k_j}, \Delta{y^k_j}, \Delta{w^k_j}, \Delta{h^k_j})$ represents the velocity, and $\Delta$ represents the change between consecutive timesteps computed as:
\begin{equation}
    \begin{array}{cc}
        \Delta{u^k_j} = u^k_j - u^k_{j-1}  & \forall u \in \{x,y, w, h\}. 
    \end{array}
    \label{deltaU}
\end{equation}

As shown in Figure \ref{fig:jlaforecast}, an RNN (PastEncoder) takes the sequence of past bounding boxes and velocities, $\boldsymbol{B}$, and generates a final hidden state vector $h^e_p$, that summarizes the sequence. The final hidden state vector is achieved by repeatedly updating the previous hidden state vector with the input $\{B^k_j\}$ for $p$ timesteps. The hidden state vector is initialized to zero. The resulting final hidden state vector is then passed through a fully connected layer with ReLU activations to generate a 256-dimensional feature vector $\phi^e_B$. 

\subsection{Embedding Encoder} \label{sec:emb}
The embedding encoder is used to capture features from the DLA-34 backbone network. The DLA-34 network provides visual context for the predicted objects' bounding boxes in the current frame. The use of visual features in the trajectory forecasting model improves the accuracy of the bounding box estimations. The DLA-34 features are shared across the detection, re-ID, and forecast branches.

Given an input frame with dimension $H_{s} \times W_{s}$, we append a forecast head to the DLA-34 network to generate a feature map $\boldsymbol{D} \in \mathbb{R}^{256 \times H \times W}$ where $H=H_{s}/4$ and $W=W_{s} / 4$. The top $N$ features of $\boldsymbol{D}$ are selected resulting in  a $256 \times N$ embedding, where $N$ is the maximum number of objects for multi-scale learning and set to a default value $N=500$. This allows us to learn a high-dimensional vector representation of the input frame in the forecasting network. 

We then pass the feature map $\boldsymbol{D}$ to a fully connected layer to generate a $256$-dimensional vector $\phi^e_E$. The resulting encoding is concatenated to the past bounding box and velocity encoding $\phi^e_B$, similar to STED \cite{Styles2020MultipleEnvironments}. However, STED uses optical flow to pass visual information to the forecasting network instead of the DLA-34 feature embeddings. The ablation study (Table \ref{tab:emb}) shows that using DLA-34 features embeddings in the forecasting model helps to improve the performance of the entire JLA. 

\subsection{Past Bounding Box and Velocity Decoder}
Another RNN (PastDecoder in Figure \ref{fig:jlaforecast}) is used to reproduce the past bounding boxes and velocities. The purpose of using the decoder to reproduce the input is to ensure that the model learns the correct input representation \cite{Ansari2020SimpleCpu}.  Thus, we can define an objective function to penalize the decoder when it deviates from the input.

At each timestep, the decoder first uses the encoding $\phi^e_B$ to update a previous hidden state vector and then passes the updated hidden state vector through a fully connected layer to generate an 8-dimensional vector. The hidden state vector is initialized to the final hidden state vector $h^e_p$ of the encoder. The output of the decoder is a set of  predicted past bounding boxes and velocities. Similar to the ground truth bounding boxes and velocities, $\boldsymbol{B}$, the predicted past bounding boxes and velocities can be represented as a sequence of $p$ sets of 8-dimensional vectors $\boldsymbol{\hat{B}} \in \mathbb{R}^{p \times 8}$ where $\boldsymbol{\hat{B}} \equiv \{\{\hat{B}^k_j\}^K_{k=1}\}^{s-1}_{j=s-p} \equiv \{\{\hat{B}^k_{s-p}\}, \{\hat{B}^k_{s-p+1}\}, \ldots, \{\hat{B}^k_{s-1}\}\}$ and $\hat{B}^k_j=(\hat{x}^k_j, \hat{y}^k_j, \hat{w}^k_j, \hat{h}^k_j, \Delta{\hat{x}^k_j}, \Delta{\hat{y}^k_j}, \Delta{\hat{w}^k_j}, \Delta{\hat{h}^k_j})$.
We use an L1 loss to penalize the decoder as follows:
\begin{equation}
\label{L1Loss}
    L_{past} = \frac{1}{(K \times p \times 8)}\sum^K_{k=1}\sum^{s-1}_{j={s-p}}{\|B^k_j - \hat{B}^k_j\|_1}.
\end{equation}

\subsection{Future Velocity Decoder}
A third RNN (FutureDecoder in Figure \ref{fig:jlaforecast}) is used to predict $q$ future velocities for the identified $K$ objects. The past bounding boxes encoding $\phi^e_B$ is concatenated with the embedding $\phi^e_E$, resulting in a 512-dimensional vector $\phi_C$. As shown in Figure \ref{fig:jlaforecast}, $\phi_C$ is passed through a fully connected ReLU layer $q$ times while updating the previous hidden state vector at each timestep. The hidden state vector is initialized to the final hidden state vector $h^e_p$ of the encoder. The decoder generates predicted future velocities $\boldsymbol{\hat{V}} \in \mathbb{R}^{q \times 4}$ where $\boldsymbol{\hat{V}} \equiv \{\{\hat{V}^k_j\}^K_{k=1}\}^{s+q}_{j=s} \equiv \{\{\hat{V}^k_{s}\}, \{\hat{V}^k_{s+1}\}, \ldots, \{\hat{V}^k_{s+q}\}\}$ and $\hat{V}^k_j = (\Delta{\hat{x}^k_j}, \Delta{\hat{y}^k_j}, \Delta{\hat{w}^k_j}, \Delta{\hat{h}^k_j})$. We do not penalize this decoder directly based on recommendations in \cite{Ansari2020SimpleCpu}. The trajectory concatenation layer, discussed next, is used to penalize the future bounding boxes instead.


\subsection{Trajectory Concatenation Layer}
The trajectory concatenation layer is used to transform the future velocities to bounding boxes \cite{Ansari2020SimpleCpu}. This layer adds the last frame bounding boxes $\{b^k_{s-1}\}^K_{k=1}$, to the cumulative sum of the velocities to generate the predicted future bounding boxes $\hat{\boldsymbol{F}} \in \mathbb{R}^{q \times 4}$, where $\hat{\boldsymbol{F}} \equiv \{\{\hat{b}^k_j\}^K_{k=1}\}^{s+q}_{j=s} \equiv \{\{\hat{b}^k_{s}\}, \{\hat{b}^k_{s+1}\}, \{\hat{b}^k_{s+2}\}, \ldots, \{\hat{b}^k_{s+q}\}\}$. The cumulative sum of the velocities is computed using Eq. \ref{eq:cumsum}. The cumulative sum of the velocities is used to compute the predicted future bounding boxes (Eq. \ref{eq:futureBB}).
\begin{equation}
\label{eq:cumsum}
 \hat{Z}^k_{i} =
\begin{cases} 
    \hat{V}^k_i & \text{if } i = 1 \vspace{2mm}\\
    \hat{V}^k_i + \hat{Z}^k_{i-1} & \text{for } i=2\dots q
\end{cases}
\end{equation}
\begin{equation}
\label{eq:futureBB}
\begin{array}{cc}
    \hat{b}^k_{s+i-1} = b^k_{s-1} + (i \times \hat{Z}^k_i) & \text{for } i = 1 \dots q.
\end{array}
\end{equation}
Hence, we can define an objective function to penalize the predicted future locations. 
We use an L1 loss function defined as:
\begin{equation}
    L_{future} = \frac{1}{(K \times q \times 4)}\sum^K_{k=1}\sum^{s+q}_{j=s}{\|\hat{b}^k_j - \hat{b}^k_j\|_1}.
\end{equation}
We compute the forecast loss as:
\begin{equation}
    L_{for} = L_{past} + L_{future}.
\end{equation}

\section{Training JLA}
The entire network is trained end-to-end using a multi-task uncertainty loss \cite{Cipolla2018Multi-taskSemantics}. The multi-task uncertainty loss performs a weighted linear sum of the losses for each task and the weights are learned automatically from the data \cite{Cipolla2018Multi-taskSemantics}. Given the detection loss $L_{det}$, re-ID loss $L_{id}$, and forecast loss $L_{for}$, the total loss function is defined as: 
\begin{equation}
\begin{array}{c}
L_{total} = \frac{1}{2}(
    {e^{-s_{det}}}L_{det} + 
    {e^{-s_{id}}}L_{id} + 
    {e^{-s_{for}}}L_{for}  
   \\
    \qquad {} + s_{det} + s_{id} + s_{for}),
\end{array}
\end{equation}
where $s_{det}$, $s_{id}$ and $s_{for}$ are the weights for detection, re-ID, and forecast, respectively. We initialize the weights to values between -2.0 to 5.0 \cite{Cipolla2018Multi-taskSemantics}, which are then  updated automatically by the model. The equations for $L_{det}$ and $L_{id}$ are defined in FairMOT \cite{Zhang2020FairMOT:Tracking}. 

The input to the model is the current frame and the past bounding boxes for the observed objects in previous frames. We use a Gated Recurrent Unit (GRU) for the PastEncoder, PastDecoder, and FutureDecoder implementation. However, any other type of RNN can be used for the implementation of the encoder and decoders.
Although the past bounding boxes are supplied for the trajectory forecasting branch, we observe that the performance of the two other branches, \ie, the detection and re-ID branches, improve greatly due to the shared image embedding (Section \ref{sec:exp}). 

Different from existing trajectory forecasting methods, we propose using a variable length of past and future bounding boxes during training and a fixed length for evaluation. This implies that each object can have any number of past and future bounding boxes, less than or equal to the fixed values of $p$ and $q$, respectively. This allows the network to learn the  objects' historical information early during training without waiting for a complete set of past or future bounding boxes. Based on this idea, we re-formalize the past and future losses as:
\begin{equation}
    L_{past} = \frac{1}{(\sum^K_{k=1}{p^k_s \times 8})}\sum^K_{k=1}\sum^{s-1}_{j={s-p^k_s}}{\|B^k_j - \hat{B}^k_j\|_1} ,
\end{equation}
\begin{equation}
    L_{future} = \frac{1}{( \sum^K_{k=1}{q^k_s \times 4})}\sum^K_{k=1}\sum^{s+q^k_s}_{j=s}{\|b^k_j - \hat{b}^k_j\|_1}.
\end{equation}
where $p^k_s \leq p$ and $q^k_s \leq q$ represents the number of ground truth past and future bounding boxes available at frame $f_s$ for each tracked object $k$, respectively. During training, we only use $p^k_s$ past and $q^k_s$ future bounding boxes for computing the loss despite predicting $p$ past and $q$ future bounding boxes.

\section{Online Inference}
In this section, we present the network inference and data association for JLA. The online association is similar to that used by FairMOT \cite{Zhang2020FairMOT:Tracking}, however, we replace the Kalman Filter estimations with trajectory forecasts and add an extra step for estimating objects' location during occlusion. The algorithm for this step is explained in Section \ref{sec:alg2}.

\subsection{Network Inference}
Our complete JLA network predicts trajectory forecasts in addition to the heatmap, bounding box offset, and bounding box size generated by the base model \cite{Zhang2020FairMOT:Tracking}. 
The size of the input frame is $1088 \times 608$,  as in previous literature \cite{Wang2020TowardsTracking, Zhang2020FairMOT:Tracking}. We initialize the past bounding box and velocity information to zero in the first three frames. As new objects are detected and tracked, we store their previous locations. We require at least two previous locations to predict future locations for an object. 

\subsection{Data Association} \label{sec: assoc}
We use the detections and trajectory forecasts for data association. The trajectory forecasts serve two purposes at inference time: (i) to generate short term estimations used in lieu of a Kalman Filter to prevent associating detections with large motion
and (ii) to generate bounding box estimation when an object is occluded.

In the first frame, all detections above the confidence threshold are initialized as new tracks. Detections in subsequent frames are linked using three key steps discussed in the next subsections. We set the state of tracks that are unmatched after the three steps to lost. If a track is lost for a predetermined number of time, we remove the track. We set the maximum lost time to $30$ frames.

\begin{algorithm} [t]
\small
\caption{SHORT-TERM FORECAST FOR MOTION FUSION}
\label{algo:motion}
\begin{algorithmic}
\Procedure{fuseMotion}{$\textit{reidDist}, \textit{tracks}, \textit{detections}, \lambda, l$}
     \For {$i \gets 0$ to $\textit{length}(\textit{tracks})$}
        \State {$\textit{track} \gets \textit{tracks}[i]$}
        \State {$d \gets \textit{IOUDistance}(\textit{track}, \textit{detections})$}
        \If{$\textit{hasForecasts}(\textit{track}) $}
            \State{$\textit{forecasts} \gets \textit{getForecasts}(\textit{track}, l)$}
            \State {$\textit{dists} \gets \textit{IOUDistance}(\textit{forecasts}, \textit{detections})$}
            \State {$m \gets \textit{minimum}(\textit{dists}, \textit{axis}=0)$}
            \State {$d \gets d * m$}
        \EndIf
        \State {$\textit{reidDist}[i, d\geq1] \gets \textit{reidDist}[i, d\geq1] * 2$}
        \State {$ \textit{costs}[i] \gets \lambda * \textit{reidDist}[i] + (1-\lambda) * d $}
    \EndFor
    \State \Return \textit{costs}
\EndProcedure
\end{algorithmic}
\end{algorithm}

\vspace{1mm}\noindent\textbf{Re-ID Features and Motion Fusion.} \label{sec:alg1}
The purpose of this step is to match tracked objects with new detections using re-ID features and bounding box overlap. A cosine distance matrix between tracks and detections computed on re-ID features is fused with short term motion distance computed on trajectory forecasts. 
The motion distance is estimated using the IOU distance between a subset of the predicted trajectory forecasts and the detections. The trajectory forecast with the minimum IOU distance is selected and used to refine the re-ID distance matrix (Algorithm \ref{algo:motion}). If the minimum IOU distance between each detection and the predicted trajectory forecasts is too large, we increase its re-ID cosine distance by a factor of two. This helps to mitigate identity switches caused by incorrect re-ID features. A cost matrix is computed by calculating a weighted sum of the re-ID cosine distance and the minimum IOU distance (Algorithm \ref{algo:motion}). In Algorithm \ref{algo:motion}, $\lambda$ is a weight to regularize the re-ID and IOU distances and $l$ is the number of trajectory forecasts used to compute the IOU distance. We set $\lambda=0.75$ and $l=10$. 
Matches are found by passing the cost matrix to an Hungarian method \cite{Kuhn1955TheProblem} for linear assignment. The ablation study in Section \ref{sec:ablation} shows that the trajectory forecasts are more accurate than a Kalman Filter for short-term motion predictions.

\vspace{1mm}\noindent\textbf{IOU Association.}
We use the IOU distance to associate unmatched tracks and unmatched detections in situations where the re-ID features and short-term forecasts are not sufficient for data association. Matches are found by using a Hungarian method for linear assignment \cite{Kuhn1955TheProblem} on the IOU distance.

\vspace{1mm}\noindent\textbf{Forecast Association.} \label{sec:alg2}
We introduce a new step for data association in which trajectory forecasts are used to estimate objects location during occlusion. At this stage, the unmatched tracks are either false positives or occluded. When an object is occluded, visual information to re-identify the object is not available. Trajectory forecasts can provide spatio-temporal information to estimate an object’s location during occlusion. We compute a cost using the distance of the trajectory forecast to the centre of the frame and the lost time of the track (Algorithm \ref{algo:forecast}). We assume that an unmatched track at the centre of the frame is likely to be occluded. The lost time of the track is incremented when a track is not associated with a new detection. The lost time prevents keeping an undetected object alive infinitely. We set $\lambda=0.5$, $\textit{maxTime}=20$, and $\textit{thresh}=0.55$ in Algorithm \ref{algo:forecast}.


\begin{algorithm} [t]
\small
\caption{FORECAST FOR OCCLUSION} \label{algo:forecast}
\begin{algorithmic}
\Procedure{forecast}{$\textit{track}, \textit{frameCenter}, \lambda, \textit{maxTime}, \textit{thresh}$}
    \If{$\textit{hasForecasts}(\textit{track}) $}
        \State{$\textit{cost} \gets \textit{track}.\textit{lostTime}/maxTime$}
        \State {$\textit{forecast} \gets \textit{getForecastAtLostTime}(\textit{track})$}
        \State {$ \textit{dist} \gets \textit{frameDistance}(\textit{forecast}, \textit{frameCenter})$}
        \State {$ \textit{cost} \gets \lambda * \textit{dist} + (1-\lambda) * \textit{cost} $}
        \If{$\textit{cost} < \textit{thresh}$}
            \State \Return  \textit{forecast}
        \EndIf
    \EndIf
    \State \Return \textit{none}
\EndProcedure
\end{algorithmic}
\end{algorithm}



\setlength{\tabcolsep}{4pt} 
\begin{table*}[t]
\begin{center}
    \begin{tabular}{l|cccc|cccc}
        \hline
          	  &  IDF1 $\uparrow$ & MT $\uparrow$ & IDs $\downarrow$ & MOTA $\uparrow$ & AIOU $\uparrow$    & FIOU $\uparrow$     &   ADE $\downarrow$ &   FDE $\downarrow$    \\
         \hline
         FairMOT &  70.9 & 140 & 441 & 67.1 & - & - & - & - \\
         FairMOT\_CV &  72.8 & 156 & 357 & 67.5 & 33.7 &   14.9 &   112.3 &  205.1 \\
         FairMOT\_KF &  72.5 & 149 & 279 & 68.1 & 38.1 &   18.9 &   104.9 &  195.3 \\
         JLA & \textbf{75.3} & \textbf{169} &\textbf{262} & \textbf{69.1} & \textbf{39.1} & \textbf{22.1} &   \textbf{97.0} &  \textbf{177.5} \\
         \hline
        
    \end{tabular}
    \end{center}
    \caption{Comparison of baseline methods with JLA on the MOT17 dataset. We set the number of past bounding boxes used for prediction to $p=10$ and number of predictions to $q=60$.}
    \label{tab:kf}
\end{table*}

\section{Experiments} \label{sec:exp}
\subsection{Datasets} \label{sub: datasets}

We evaluate the performance of JLA using different amount of training data. For ablation studies (Section \ref{sec:ablation}), we train JLA on $50\%$ of MOT17 training set and evaluate the model on the remaining $50\%$. 
For evaluation on the MOTChallenge server, we use the same training data as FairMOT \cite{Zhang2020FairMOT:Tracking}. The training data is described below:
\begin{itemize}
    \item We use ETH \cite{Ess2008ATracking} and CityPersons \cite{Zhang2017CityPersons:Detection} to train the detection branch because these datasets provide only bounding box annotations.
    \item We use PRW \cite{Zheng2017PersonTheWild}, MOT17 \cite{Milan2016MOT16:Tracking}, and CalTech \cite{Dollar2010PedestrianBenchmark} to train the detection, re-ID, and trajectory forecast branches because these datasets provide both bounding box and identity annotations. 
    \item We use CUHK-SYS \cite{Xiao2017JointSearch} to train the detection and re-ID branches. CUHK-SYS provides both bounding box and identity annotations but the frames are not sequential and is therefore not suitable for trajectory forecasting.  
\end{itemize}
We evaluate the tracking result of JLA on the tests sets of the MOT15, MOT16, MOT17, and MOT20 benchmarks (Section \ref{sec:result}).

\subsection{Implementation details}
We finetune JLA on a model pre-trained on the CrowdHuman dataset \cite{Shao2018CrowdHuman:Crowd}. The pre-training process uses a self-supervised approach in which a unique identity label is assigned to each bounding box and the model is trained without any tracking information \cite{Zhang2020FairMOT:Tracking}.  

We train JLA on six datasets (Section \ref{sub: datasets}) with Adam optimizer for 30 epochs with a starting learning rate of ${10}^{-4}$, which decays to ${10}^{-5}$ after 20 epochs. The training takes about 60 hours on two GeForce RTX 2080 GPUs with a batch size of 8. We use the standard data augmentation techniques used in literature including color jittering, rotation, and scaling. Finally, we finetune JLA for 20 epochs on the training datasets of MOT20 and MOT15 to evaluate their respective test datasets. The results on the MOTChallenge are discussed in Section \ref{sec:result}.

\setlength{\tabcolsep}{4pt} 
\begin{table}[]
\begin{center}
    \begin{tabular}{ccc|cccc}
        \hline
           \makecell{App + \\ Forecast} &   
           \makecell{Box\\ IOU} &
           \makecell{Occlusion \\ Forecast}  &
           \makecell{IDF1 \\ $\uparrow$} &
           \makecell{MT \\ $\uparrow$} &
           \makecell{IDs \\ $\downarrow$} &
           \makecell{MOTA \\ $\uparrow$} \\
         \hline 
         \cmark   &  \xmark & \xmark   & 72.7 & 139 &   381 &  67.7  \\
         \cmark   & \cmark  & \xmark    & 72.8 & 139 &  366 & 67.8\\
        \xmark & \cmark   &\xmark & 60.4  &  145 & 1185 &  64.3  \\
         \xmark & \cmark  & \cmark  & 65.0 & 145 & 976 & 64.8\\
         \cmark   & \xmark & \cmark  & 74.7 &  168 &301 & 68.9\\
         \cmark & \cmark  & \cmark  &  \textbf{75.3} &  \textbf{169} &  \textbf{262} & \textbf{69.1}\\
         \hline
    \end{tabular}
    \end{center}
    \caption{Evaluation of the data association components in JLA on the MOT17 dataset.  We set the number of past bounding boxes used for prediction to $p=10$ and number of predictions to $q=60$.}
    \label{tab:comp}
\end{table}

\setlength{\tabcolsep}{4pt} 
\begin{table}[]
\begin{center}
    \begin{tabular}{c|cccc}
        \hline
          	 DLA34 Embedding &  IDF1 $\uparrow$  & MT $\uparrow$    &   IDs $\downarrow$  &   MOTA $\uparrow$\\
         \hline
         \xmark &   72.8 & 150 & \textbf{262} & 65.8 \\
         \cmark &  \textbf{75.3} &  \textbf{169} & \textbf{262} & \textbf{69.1} \\
         \hline
        
    \end{tabular}
    \end{center}
    \caption{Evaluation of DLA34 feature embedding in JLA on the MOT17 dataset. We set the number of past bounding boxes used for prediction to $p=10$ and number of predictions to $q=60$.}
    \label{tab:emb}
\end{table}

\setlength{\tabcolsep}{4pt} 
\begin{table*}[t]
\begin{center}
    \begin{tabular}{l|l|llllll}
        \hline
          	 Dataset &  Tracker &   MOTA $\uparrow$    & IDF1 $\uparrow$      & MT $\uparrow$    & ML $\downarrow$     &   IDs $\downarrow$   & FPS $\uparrow$\\
         \hline
         
         \multirow{1}{*}{MOT15}  &  TubeTK \cite{Pang_2020_CVPR}   &   58.4   &  53.1   &    39.3   &    18.0   &    854   &    5.8  \\
         &  FairMOT \cite{Zhang2020FairMOT:Tracking}&  \textbf{60.6}   &    \textbf{64.7 }  &    \textbf{47.6}   &    \textbf{11.0}  &    \textbf{591}   &   \textbf{30.5}    \\
         &  JLA (Ours)   &  55.8   & 63.2  &   42.7   &   16.1   & 644   &  22.4 \\
         \hline
         
         \multirow{1}{*}{MOT16} 
         &  TubeTK \cite{Pang_2020_CVPR}   &   64.0   &  59.4   &    33.5   &    19.4   &    1117   &    1.0  \\
         & JDE \cite{Wang2020TowardsTracking}  &  64.4 &  55.8 &  35.4 & 20.0  & 1544  & 18.5\\
         &  CTrackerV1 \cite{Peng2020Chained-Tracker:Tracking}   &    67.6   & 57.2  &    32.9   &    23.1   &    1897   &    6.8 \\
         &  FairMOT \cite{Zhang2020FairMOT:Tracking}&  \textbf{74.9}   &    72.8   &    44.7   &    \textbf{15.9 }  &    1074   &   \textbf{25.9}    \\
         
          &  JLA (Ours)   &  73.8   &   \textbf{75.0}  &   \textbf{44.9 }  &   22.8   & \textbf{719}   &  19.7 \\
         \hline

         \multirow{1}{*}{MOT17} &  TubeTK \cite{Pang_2020_CVPR}   &   63.0   &  58.6   &    31.2   &    19.9   &    4137   &    3.0  \\
         &  CTrackerV1 \cite{Peng2020Chained-Tracker:Tracking}   &    66.6   & 57.4  &    32.2   &    24.2   &    5529   &    6.8 \\
         &  FairMOT \cite{Zhang2020FairMOT:Tracking}&  73.7   &    72.3   &    43.2   &    \textbf{17.3 }  &    3303   &   \textbf{25.9}    \\
         &  JLA (Ours)   &  \textbf{74.0}   & \textbf{74.0}  &   \textbf{45.1 }  &   20.5   & \textbf{2292}   &  19.0 \\
         \hline
         
         \multirow{1}{*}{MOT20} &  FairMOT \cite{Zhang2020FairMOT:Tracking}&  \textbf{61.8 }  &    67.3   &    68.8   &    \textbf{7.6}  &    5243   &   \textbf{25.9}    \\
         &  JLA (Ours)   &  60.2   & \textbf{68.7}  &  59.8  &  10.5    & \textbf{2780 } &  14.3 \\
         \hline
        
    \end{tabular}
    \end{center}
    \caption{Comparison of state-of-the-art methods under the ``private" category on the MOTChallenge benchmarks. These methods are categorized under the \emph{private detections} because they use more datasets in training the model. }
    \label{tab:mot}
\end{table*}

\subsection{Ablation Studies} \label{sec:ablation}
We compare the performance of JLA with a constant velocity and the Kalman Filter baseline methods. We also perform ablation studies to evaluate the impact of the various components of JLA. We discuss these evaluations in the next subsections.

\vspace{1mm}\noindent\textbf{Baselines.}
A constant velocity model and the Kalman Filter are often used as baseline methods in trajectory forecasting. For a fair comparison of the JLA with the baseline methods, we modify the data association step in FairMOT \cite{Zhang2020FairMOT:Tracking} to be the same as in JLA. Then, for each detected object, we generate $q$ trajectory forecast predictions using a constant velocity (FairMOT\_CV in Table \ref{tab:kf}) and a Kalman Filter (FairMOT\_KF in Table \ref{tab:kf}). We set $q=60$. We evaluate the tracking and trajectory forecasting performance using the metrics mentioned in section \ref{sec: metrics}. We do not compare our work with STED because STED requires an external detector to precompute the trajectories, which will result in an unfair comparison.

The results in Table \ref{tab:kf} show that using our proposed data association steps improves the performance of the base method, FairMOT \cite{Zhang2020FairMOT:Tracking}. The number of ID switches (IDs) is reduced considerably compared to both cases, constant velocity and the Kalman Filter. Also, the number of mostly tracked objects and accuracy increases. This shows that we can detect some occluded objects using trajectory forecasts.

In addition, the table shows that the trajectory forecasting branch in JLA performs better than the constant velocity and Kalman Filter predictions with an increase in AIOU and FIOU, and a decrease in ADE and FDE.

Next, we study the impact of each component of JLA on the data association.



\vspace{1mm}\noindent\textbf{Analysis of Data Association Components in JLA.}
As discussed in Section \ref{sec: assoc}, JLA uses three components for data association: appearance embedding fused with short-term forecasts, bounding box IOU, and trajectory forecast predictions during occlusion. We study the impact of these individual components on the performance of the model. Where applicable, one or more components are turned off and the model is evaluated without the component(s). The results of the study are shown in Table \ref{tab:comp}. 

The best performance is achieved when all the components are turned on. When \textit{trajectory forecast during occlusion} is turned off, the number of ID switches (IDs) increases from $262$ to $366$; the number of mostly tracked decreases from $169$ to $139$; IDF1 and MOTA decrease from $75.3\%$ to $72.8\%$ and $69.1\%$ to $67.7\%$ respectively. Associating data based on bounding box IOU alone gives the worst performance.

Next, we study the effect of removing the image embedding from the trajectory forecasting network. 

\vspace{1mm}\noindent\textbf{Image Embedding in Trajectory Forecasting.}
As explained in Section \ref{sec:emb}, the image embedding from the  DLA34 network provides visual features for the trajectory forecasting network. Previous literature concatenates optical flow encoding to the previous bounding box encoding \cite{Styles2020MultipleEnvironments} to provide visual information to the trajectory forecasting network. Optical flow is computationally expensive and requires to be computed separately  \cite{Styles2020MultipleEnvironments}. In our work, we concatenate the DLA34 output features with the previous bounding box encoding. This approach is less expensive and provides high dimensional features for the trajectory forecasting network. 
We train JLA without this image embedding and compare its performance against the case of JLA trained with the embedding.

The result in Table \ref{tab:emb} shows that including DLA34 features in the trajectory forecasting network improves the performance of JLA. MOTA increases from $65.8\%$ to $69.1\%$, IDF1 increases from $72.8\%$ to $75.3\%$, and number of mostly tracked increases from from $150$ to $169$.



        

\subsection{Results on MOTChallenge} \label{sec:result}
We compare our method with the top methods in the MOTChallenge Benchmarks under the private category. The private category uses an external detector or datasets for training. As shown in Table \ref{tab:mot}, JLA reduces the ID switches (IDs) compared to FairMOT by $33\%$, $31\%$, and $47\%$ for MOT16, MOT17, and MOT20, respectively. JLA does not perform as good on MOT15 because the performance of the trajectory forecast is dependent on the availability of ground truth information. MOT15 has missing tracking information during occlusion, which can result in a  high number of false positives.


\section{Conclusion}
We have introduced a joint learning architecture for multiple object tracking and trajectory forecasting. We have shown that trajectory forecasting can be used in lieu of Kalman Filters to model non-linear trajectories. Also, we have shown that future predictions can be used to estimate objects' locations during occlusion. Our evaluations on the MOTChallenge benchmarks show that our architecture reduces the identity switches within an MOT context considerably. Our work shows promises for future research to develop more sophisticated architectures that improves multiple object tracking and trajectory forecasting.  

{\small
\bibliographystyle{ieee_fullname}
\bibliography{references, egbib}

\begin{thebibliography}{10}\itemsep=-1pt

\bibitem{Alahi2016SocialSpaces}
Alexandre Alahi, Kratarth Goel, Vignesh Ramanathan, Alexandre Robicquet, Li
  Fei-Fei, and Silvio Savarese.
\newblock {Social LSTM: Human trajectory prediction in crowded spaces}.
\newblock In {\em Proceedings of the IEEE Computer Society Conference on
  Computer Vision and Pattern Recognition}, volume 2016-December, pages
  961--971. IEEE Computer Society, 12 2016.

\bibitem{Ansari2020SimpleCpu}
Junaid~Ahmed Ansari and Brojeshwar Bhowmick.
\newblock {Simple means Faster: Real-time human motion forecasting in monocular
  first person videos on CPU}.
\newblock Technical report, 2020.

\bibitem{Bergmann2019TrackingWhistles}
Philipp Bergmann, Tim Meinhardt, and Laura Leal-Taixe.
\newblock {Tracking without bells and whistles}.
\newblock Technical report, 2019.

\bibitem{Bernardin2008EvaluatingMetrics}
Keni Bernardin and Rainer Stiefelhagen.
\newblock {Evaluating Multiple Object Tracking Performance: The CLEAR MOT
  Metrics}.
\newblock {\em EURASIP Journal on Image and Video Processing}, 2008(1):1--10, 5
  2008.

\bibitem{Bewley2016SimpleTracking}
Alex Bewley, Zongyuan Ge, Lionel Ott, Fabio Ramos, and Ben Upcroft.
\newblock {Simple online and realtime tracking}.
\newblock Technical report, 2016.

\bibitem{8015000}
Jiahui Chen, Hao Sheng, Yang Zhang, and Zhang Xiong.
\newblock Enhancing detection model for multiple hypothesis tracking.
\newblock In {\em 2017 IEEE Conference on Computer Vision and Pattern
  Recognition Workshops (CVPRW)}, pages 2143--2152, 2017.

\bibitem{choi2010multiple}
Wongun Choi and Silvio Savarese.
\newblock Multiple target tracking in world coordinate with single, minimally
  calibrated camera.
\newblock In {\em European Conference on Computer Vision}, pages 553--567.
  Springer, 2010.

\bibitem{Cipolla2018Multi-taskSemantics}
Roberto Cipolla, Yarin Gal, and Alex Kendall.
\newblock {Multi-task Learning Using Uncertainty to Weigh Losses for Scene
  Geometry and Semantics}.
\newblock In {\em Proceedings of the IEEE Computer Society Conference on
  Computer Vision and Pattern Recognition}, pages 7482--7491. IEEE Computer
  Society, 12 2018.

\bibitem{grid-based-goals}
Nachiket Deo and Mohan~M. Trivedi.
\newblock {Trajectory Forecasts in Unknown Environments Conditioned on
  Grid-Based Plans}.
\newblock {\em arXiv preprint arXiv:2001.00735}, 2020.

\bibitem{Dollar2010PedestrianBenchmark}
Piotr Dollar, Christian Wojek, Bernt Schiele, and Pietro Perona.
\newblock {Pedestrian detection: A benchmark}.
\newblock In {\em 2009 IEEE Conference on Computer Vision and Pattern
  Recognition, Miami, FL, 2009}, pages 304--311. Institute of Electrical and
  Electronics Engineers (IEEE), 3 2010.

\bibitem{Ess2008ATracking}
Andreas Ess, Bastian Leibe, Konrad Schindler, and Luc Van~Gool.
\newblock {A mobile vision system for robust multi-person tracking}.
\newblock In {\em 26th IEEE Conference on Computer Vision and Pattern
  Recognition, CVPR}, 2008.

\bibitem{henriques2011globally}
Joao~F Henriques, Rui Caseiro, and Jorge Batista.
\newblock Globally optimal solution to multi-object tracking with merged
  measurements.
\newblock In {\em 2011 International Conference on Computer Vision}, pages
  2470--2477. IEEE, 2011.

\bibitem{7410890}
Chanho Kim, Fuxin Li, Arridhana Ciptadi, and James~M. Rehg.
\newblock Multiple hypothesis tracking revisited.
\newblock In {\em 2015 IEEE International Conference on Computer Vision
  (ICCV)}, pages 4696--4704, 2015.

\bibitem{Kuhn1955TheProblem}
H.~W. Kuhn.
\newblock {The Hungarian method for the assignment problem}.
\newblock {\em Naval Research Logistics Quarterly}, 2(1-2):83--97, 3 1955.

\bibitem{desire}
Namhoon Lee, Wongun Choi, Paul Vernaza, Christopher~B Choy, Philip H~S Torr,
  and Manmohan Chandraker.
\newblock {Desire: Distant future prediction in dynamic scenes with interacting
  agents}.
\newblock In {\em IEEE Conference on Computer Vision and Pattern Recognition},
  2017.

\bibitem{4809540}
Young-Sook Lee and HoonJae Lee.
\newblock Multiple object tracking for fall detection in real-time surveillance
  system.
\newblock In {\em 2009 11th International Conference on Advanced Communication
  Technology}, volume~03, pages 2308--2312, 2009.

\bibitem{Lin2021OnTracking}
Xufeng Lin, Chang~Tsun Li, Victor Sanchez, and Carsten Maple.
\newblock {On the detection-to-track association for online multi-object
  tracking}.
\newblock {\em Pattern Recognition Letters}, 146:200--207, 6 2021.

\bibitem{LUO2021103448}
Wenhan Luo, Junliang Xing, Anton Milan, Xiaoqin Zhang, Wei Liu, and Tae-Kyun
  Kim.
\newblock Multiple object tracking: A literature review.
\newblock {\em Artificial Intelligence}, 293:103448, 2021.

\bibitem{kartik-endpoints}
Karttikeya Mangalam, Harshayu Girase, Shreyas Agarwal, Kuan-Hui Lee, Ehsan
  Adeli, Jitendra Malik, and Adrien Gaidon.
\newblock {It is not the journey but the destination: Endpoint conditioned
  trajectory prediction}.
\newblock In {\em European Conference on Computer Vision}, 2020.

\bibitem{Milan2016MOT16:Tracking}
Anton Milan, Laura Leal-Taixe, Ian Reid, Stefan Roth, and Konrad Schindler.
\newblock {MOT16: A Benchmark for Multi-Object Tracking}.
\newblock Technical report, 2016.

\bibitem{Pang2020Tubetk:Model}
Bo Pang, Yizhuo Li, Yifan Zhang, Muchen Li, and Cewu Lu.
\newblock {Tubetk: Adopting tubes to track multi-object in a one-step training
  model}.
\newblock Technical report, 2020.

\bibitem{Pang_2020_CVPR}
Bo Pang, Yizhuo Li, Yifan Zhang, Muchen Li, and Cewu Lu.
\newblock Tubetk: Adopting tubes to track multi-object in a one-step training
  model.
\newblock In {\em Proceedings of the IEEE/CVF Conference on Computer Vision and
  Pattern Recognition (CVPR)}, June 2020.

\bibitem{Peng2020Chained-Tracker:Tracking}
Jinlong Peng, Changan Wang, Fangbin Wan, Yang Wu, Yabiao Wang, Ying Tai,
  Chengjie Wang, Jilin Li, Feiyue Huang, and Yanwei Fu.
\newblock {Chained-Tracker: Chaining Paired Attentive Regression Results for
  End-to-End Joint Multiple-Object Detection and Tracking}.
\newblock In {\em Lecture Notes in Computer Science (including subseries
  Lecture Notes in Artificial Intelligence and Lecture Notes in
  Bioinformatics)}, volume 12349 LNCS, pages 145--161. Springer Science and
  Business Media Deutschland GmbH, 7 2020.

\bibitem{Pi2020JointlyInformation}
Zhixiong Pi, Huai Qin, Changxin Gao, and Nong Sang.
\newblock {Jointly detecting and multiple people tracking by semantic and scene
  information}.
\newblock {\em Neurocomputing}, 412:244--251, 2020.

\bibitem{qin2012improving}
Zhen Qin and Christian~R Shelton.
\newblock Improving multi-target tracking via social grouping.
\newblock In {\em 2012 IEEE Conference on Computer Vision and Pattern
  Recognition}, pages 1972--1978. IEEE, 2012.

\bibitem{Ristani2016PerformanceTracking}
Ergys Ristani, Francesco Solera, Roger Zou, Rita Cucchiara, and Carlo Tomasi.
\newblock {Performance measures and a data set for multi-target, multi-camera
  tracking}.
\newblock In {\em Lecture Notes in Computer Science (including subseries
  Lecture Notes in Artificial Intelligence and Lecture Notes in
  Bioinformatics)}, volume 9914 LNCS, pages 17--35. Springer Verlag, 2016.

\bibitem{learning-etiquette}
Alexandre Robicquet, Amir Sadeghian, Alexandre Alahi, and Silvio Savarese.
\newblock {Learning social etiquette: Human trajectory understanding in crowded
  scenes}.
\newblock In {\em European Conference on Computer Vision}, 2016.

\bibitem{trajectory-forecasting-survey}
Andrey Rudenko, Luigi Palmieri, Michael Herman, Kris~M Kitani, Dariu~M Gavrila,
  and Kai~O Arras.
\newblock {Human motion trajectory prediction: A survey}.
\newblock {\em The International Journal of Robotics Research}, 39(8):895--935,
  2020.

\bibitem{trajnet}
Amir Sadeghian, Vineet Kosaraju, Agrim Gupta, Silvio Savarese, and Alexandre
  Alahi.
\newblock {TrajNet: Towards a Benchmark for Human Trajectory Prediction}.
\newblock {\em arXiv preprint}, 2018.

\bibitem{Shao2018CrowdHuman:Crowd}
Shuai Shao, Zijian Zhao, Boxun Li, Tete Xiao, Gang Yu, Xiangyu Zhang, and Jian
  Sun.
\newblock {CrowdHuman: A Benchmark for Detecting Human in a Crowd}.
\newblock {\em arXiv}, 4 2018.

\bibitem{Styles2020MultipleEnvironments}
Olly Styles, Tanaya Guha, and Victor Sanchez.
\newblock {Multiple object forecasting: Predicting future object locations in
  diverse environments}.
\newblock Technical report, 2020.

\bibitem{Sun2021DeepTracking}
Shijie Sun, Naveed Akhtar, Huansheng Song, Ajmal Mian, and Mubarak Shah.
\newblock {Deep Affinity Network for Multiple Object Tracking}.
\newblock {\em IEEE Transactions on Pattern Analysis and Machine Intelligence},
  43(1):104--119, 1 2021.

\bibitem{Voigtlaender2019Mots:Segmentation}
Paul Voigtlaender, Michael Krause, Aljosa Osep, Jonathon Luiten, Berin
  Balachandar~Gnana Sekar, Andreas Geiger, and Bastian Leibe.
\newblock {Mots: Multi-object tracking and segmentation}.
\newblock {\em Proceedings of the IEEE Computer Society Conference on Computer
  Vision and Pattern Recognition}, 2019-June:7934--7943, 2 2019.

\bibitem{Wang2020TowardsTracking}
Zhongdao Wang, Liang Zheng, Yixuan Liu, Yali Li, and Shengjin Wang.
\newblock {Towards Real-Time Multi-Object Tracking}.
\newblock Technical report, 2020.

\bibitem{Wojke2018SimpleMetric}
Nicolai Wojke, Alex Bewley, and Dietrich Paulus.
\newblock {Simple online and realtime tracking with a deep association metric}.
\newblock Technical report, 2018.

\bibitem{Xiao2017JointSearch}
Tong Xiao, Shuang Li, Bochao Wang, Liang Lin, and Xiaogang Wang.
\newblock {Joint detection and identification feature learning for person
  search}.
\newblock In {\em Proceedings - 30th IEEE Conference on Computer Vision and
  Pattern Recognition, CVPR 2017}, volume 2017-January, pages 3376--3385.
  Institute of Electrical and Electronics Engineers Inc., 11 2017.

\bibitem{fpl}
Takuma Yagi, Karttikeya Mangalam, Ryo Yonetani, and Yoichi Sato.
\newblock {Future Person Localization in First-Person Videos}.
\newblock In {\em Proceedings of the IEEE Computer Society Conference on
  Computer Vision and Pattern Recognition}, pages 7593--7602, 2018.

\bibitem{yang2011learning}
Bo Yang, Chang Huang, and Ram Nevatia.
\newblock Learning affinities and dependencies for multi-target tracking using
  a crf model.
\newblock In {\em CVPR 2011}, pages 1233--1240. IEEE, 2011.

\bibitem{brian-egocentric}
Yu Yao, Mingze Xu, Chiho Choi, David~J. Crandall, Ella~M. Atkins, and Behzad
  Dariush.
\newblock {Egocentric vision-based future vehicle localization for intelligent
  driving assistance systems}.
\newblock In {\em Proceedings - IEEE International Conference on Robotics and
  Automation}, volume 2019-May, pages 9711--9717, 2019.

\bibitem{Zhang2017CityPersons:Detection}
Shanshan Zhang, Rodrigo Benenson, and Bernt Schiele.
\newblock {CityPersons: A diverse dataset for pedestrian detection}.
\newblock In {\em Proceedings - 30th IEEE Conference on Computer Vision and
  Pattern Recognition, CVPR 2017}, volume 2017-Janua, pages 4457--4465.
  Institute of Electrical and Electronics Engineers Inc., 11 2017.

\bibitem{Zhang2020}
Yang Zhang, Hao Sheng, Yubin Wu, Shuai Wang, Weifeng Lyu, Wei Ke, and Zhang
  Xiong.
\newblock {Long-Term Tracking with Deep Tracklet Association}.
\newblock {\em IEEE Transactions on Image Processing}, 29:6694--6706, 2020.

\bibitem{Zhang2020FairMOT:Tracking}
Yifu Zhang, Chunyu Wang, Xinggang Wang, Wenjun Zeng, and Wenyu Liu.
\newblock {FairMOT: On the Fairness of Detection and Re-Identification in
  Multiple Object Tracking}.
\newblock Technical report, 2020.

\bibitem{Zheng2017PersonTheWild}
Liang Zheng, Hengheng Zhang, Shaoyan Sun, Manmohan Chandraker, Yi Yang, and Qi
  Tian.
\newblock {Person re-identification in theWild}.
\newblock In {\em Proceedings - 30th IEEE Conference on Computer Vision and
  Pattern Recognition, CVPR 2017}, volume 2017-Janua, pages 3346--3355.
  Institute of Electrical and Electronics Engineers Inc., 11 2017.

\end{thebibliography}
}



\end{document}